\documentclass{article}
\usepackage{spconf,amsmath,graphicx,amsfonts}
\usepackage{multirow}
\usepackage{subfig}
\usepackage{url}
\usepackage[table]{xcolor}

\title{End-to-end Automatic Speech Translation of Audiobooks}
\name{Alexandre B\'erard$^{\dagger\star}$ \qquad  Laurent Besacier$^{\star}$ \qquad Ali Can Kocabiyikoglu$^{\star}$
\qquad Olivier Pietquin$^{\dagger}$}
\address{$^{\star}$LIG - Univ. Grenoble-Alpes (France) \qquad $^{\dagger}$CRIStAL - Univ. Lille (France)}
\begin{document}

\maketitle

\begin{abstract}
We investigate end-to-end speech-to-text translation on a corpus of audiobooks specifically augmented for this task.
Previous works investigated the extreme case where source language transcription is not available during learning nor decoding, but we also study a midway case where source language transcription is available at training time only. In this case, a single model is trained to decode source speech into target text in a single pass. Experimental results show that it is possible to train compact and efficient end-to-end speech translation models in this setup. We also distribute the corpus and hope that our speech translation baseline on this corpus will be challenged in the future.
\end{abstract}


\section{Introduction}
\label{sec:intro}

Most spoken language translation (SLT)  systems integrate (loosely or closely) two main modules: source language speech recognition (ASR) and source-to-target
text translation (MT).  In these approaches, a symbolic sequence of words (or characters) in the source language is used as an intermediary representation during the speech translation process.
However, recent works have attempted to build end-to-end speech-to-text translation without using source language transcription during learning or decoding.  One attempt to translate directly a source speech signal into target language text is that of \cite{Duong2016}. However, the authors focus on the alignment between source speech utterances and their text translation without proposing a complete end-to-end translation system. The first attempt to build an end-to-end speech-to-text translation system (which does not use source language) is our own work \cite{Berard2016} but it was applied to a synthetic (TTS) speech corpus. A similar approach was then proposed and evaluated on a real speech corpus by \cite{Weiss2017}.   

This paper is a follow-up of our previous work \cite{Berard2016}. We now investigate end-to-end speech-to-text translation on a corpus of audiobooks - \textit{LibriSpeech} \cite{Panayotov2015} -  specifically augmented to perform end-to-end speech translation \cite{Alican2018}. 
While previous works \cite{Berard2016,Weiss2017} investigated the extreme case where source language transcription is not available during learning nor decoding (unwritten language scenario defined in \cite{Adda2016,Anastasopoulos2017}), we also investigate, in this paper, a midway case where a certain amount of source language transcription is available during training. In this intermediate scenario, a unique (end-to-end) model is trained to decode source speech into target text through a single pass (which can be interesting if compact speech translation models are needed).

This paper is organized as follows: after presenting our corpus in section \ref{corpus},  we present our end-to-end models in section \ref{models}. Section \ref{exp} describes our evaluation on two datasets: the synthetic dataset used in \cite{Berard2016} and the audiobook dataset described in section \ref{corpus}. Finally, section \ref{concl} concludes this work.

\section{Audiobook Corpus for End-to-End Speech Translation}
\label{corpus}

\subsection{Augmented LibriSpeech }

Large quantities of parallel texts (e.g. Europarl or OpenSubtitles) are available for training text machine translation systems, but there are no large (\textgreater 100h) and publicly available parallel corpora that include speech in a source language aligned to text in a target language.
The \textit{Fisher/Callhome} Spanish-English corpora \cite{Post2013} are only medium size (38h), contain low-bandwidth recordings, and are not available for free.

We very recently built a large English to French corpus for direct speech translation training and evaluation \cite{Alican2018}\footnote{The Augmented LibriSpeech corpus is available for download here: \url{https://persyval-platform.univ-grenoble-alpes.fr/DS91/detaildataset}}, which is much larger than the existing corpora described above. We started from the \textit{LibriSpeech} corpus used for Automatic Speech Recognition (ASR), which has 1000 hours of speech aligned with their transcriptions \cite{Panayotov2015}.

The read audiobook recordings derive from a project based on a collaborative effort: LibriVox. The speech recordings are based on public domain books available on \textit{Gutenberg Project}\footnote{\url{https://www.gutenberg.org/}} which are distributed in \textit{LibriSpeech} along with the recordings.

Our augmentation of \textit{LibriSpeech} is straightforward: we automatically aligned e-books in a foreign language (French) with English utterances of \textit{LibriSpeech}. This lead to 236 hours of English speech aligned to French translations at utterance level (more details can be found in \cite{Alican2018}). Since English (source) transcriptions are initially available for \textit{LibriSpeech}, we also translated them using \textit{Google Translate}. To summarize, for each utterance of our 236h corpus, the following quadruplet is available: English speech signal, English transcription, French text translation 1 (from alignment of e-books) and translation 2 (from MT of English transcripts).

\begin{table*}[t]
    \centering
    \begin{tabular}{|c|c|c|c||c|c|c||c|}
        \hline
         \multicolumn{2}{|c|}{\multirow{2}{*}{Corpus}} & \multicolumn{2}{c||}{Total} & \multicolumn{3}{c||}{Source (per segment)} & Target (per segment) \\
         \cline{3-8}
         \multicolumn{2}{|l|}{} & segments & hours & frames & chars & (sub)words & chars \\
         \hline
         \hline
         \multirow{4}{*}{LibriSpeech (Real)} & train 1 & \multirow{2}{*}{47271} & \multirow{2}{*}{100:00} & \multirow{2}{*}{762} & \multirow{2}{*}{111} & \multirow{2}{*}{20.7} & 143 \\
         & train 2 & & & & & & 126 \\
         \cline {2-8}
         & dev & 1071 & 2:00 & 673 & 93 & 17.9 & 110 \\
         & test & 2048 & 3:44 & 657 & 95 & 18.3 & 112 \\
         \hline
         \hline
         \multirow{3}{*}{BTEC (Synthetic)} & train & 19972 & 15:51 & 276 & 50 & 10 & 42 \\
         & dev & 1512 & 0:59 & 236 & 40 & 8.1 & 33 \\
         & test & 933 & 0:36 & 236 & 41 & 8.2 & 34 \\
         \hline
    \end{tabular}
    \caption{Size of the Augmented LibriSpeech and BTEC corpora, with the average frame, character and word counts (subword count for LibriSpeech) per segment.
    Character counts take whitespaces into account. The source side of BTEC actually has six times this number of segments and hours, because we concatenate multiple speakers (synthetic voices). LibriSpeech \emph{train 1} (alignments) and \emph{train 2} (automatic translation) share the same source side.}
    \label{tab:corpus_size}
\end{table*}

\subsection{MT and AST tasks}

This paper focuses on the speech translation (AST) task of audiobooks from English to French, using the Augmented LibriSpeech corpus.
We compare a direct (end-to-end) approach, with a cascaded approach that combines a neural speech transcription (ASR) model with a neural machine translation model (MT). The ASR and MT results are also reported as baselines for future uses of this corpus.

Augmented LibriSpeech contains 236 hours of speech in total, which is split into 4 parts: a test set of 4 hours, a dev set of 2 hours, a clean train set of 100 hours, and an extended train set with the remaining 130 hours. Table~\ref{tab:corpus_size} gives detailed information about the size of each corpus.
All segments in the corpus were sorted according to their alignment confidence scores, as produced by the alignment software used by the authors of the corpus \cite{Alican2018}.
The test, dev and train sets correspond to the highest rated alignments. The remaining data (extended train) is more noisy, as it contains more incorrect alignments. The test set was manually checked, and incorrect alignments were removed. We perform all our experiments using \emph{train} only (without \emph{extended train}). Furthermore, we double the training size by concatenating the aligned references with the Google Translate references.
We also mirror our experiments on the BTEC synthetic speech corpus, as a follow-up to \cite{Berard2016}.

\section{End-to-End Models}
\label{models}

For the three tasks, we use encoder-decoder models with attention \cite{Bahdanau2015,Chorowski2015,Bahdanau2016,Berard2016,Weiss2017}. Because we want to share some parts of the model between tasks (multi-task training), the ASR and AST models use the same encoder architecture, and the AST and MT models use the same decoder architecture.
\newcommand{\real}{\mathbb{R}}
\newcommand{\cev}[1]{\reflectbox{\ensuremath{\vec{\reflectbox{\ensuremath{#1}}}}}}

\subsection{Speech Encoder}

The speech encoder is a mix between the convolutional encoder presented in \cite{Weiss2017} and our previously proposed encoder \cite{Berard2016}. It takes as input a sequence of audio features: $\mathbf{x}=(x_1,\ldots,x_{T_x})\in\real^{T_x \times n}$.
Like \cite{Berard2016}, these features are given as input to two non-linear ($tanh$) layers, which output new features of size $n'$.
Like \cite{Weiss2017}, this new set of features is then passed to a stack of two convolutional layers. Each layer applies 16 convolution filters of shape $(3, 3, depth)$ with a stride of $(2, 2)$ w.r.t. time and feature dimensions; $depth$ is $1$ for the first layer, and $16$ for the second layer.
We get features of shape $(T_x/2,n'/2,16)$ after the 1\textsuperscript{st} layer, and $(T_x/4,n'/4,16)$ after the 2\textsuperscript{nd} layer. This latter tensor is flattened with shape $(T'_x=T_x/4,4n')$ before being passed to a stack of three bidirectional LSTMs.
This set of features has $1/4th$ the time length of the initial features, which speeds up training, as the complexity of the model is quadratic with respect to the source length.
In our models, we use $n'=128$, which gives features of size $512$.

The last bidirectional LSTM layer computes a sequence of annotations $\mathbf{h}=h_1,\cdots,h_{T'_x}$, where each annotation $h_i$ is a concatenation of the corresponding forward and backward states: $h_i=(\vec{h_i}\oplus\cev{h_i})\in\real^{2m}$, with $m$ the encoder cell size.

This model differs from \cite{Berard2016}, which did not use convolutions, but time pooling between each LSTM layer, resulting in a shorter sequence (pyramidal encoder).
\subsection{Character-level decoder}

We use a character-level decoder composed of a conditional LSTM \cite{Sennrich2017}, followed by a dense layer.
\begin{align}
    s_t,o_t=update^1(s'_{t-1},E(y_{t-1})) \\
    c_t = look(o_t,\mathbf{h}) \\
    s'_t,o'_t=update^2(s_{t-1},c_t) \\
    y_t = generate(o't\oplus c_t \oplus E(y_{t-1}))
\end{align}

where $update^1$ and $update^2$ are two LSTMs with cell size $m'$. $look$ is a vanilla global attention mechanism \cite{Bahdanau2015}, which uses a feed-forward network with one hidden layer of size $m'$. $E^{k\times |V|}$ is the target embedding matrix, with $k$ the embedding size and $|V|$ the vocabulary size. $c_t\in\real^{2m}$ is a context vector which summarizes the input states to help the decoder generate a new symbol and update its state.
\noindent $generate$ uses a non-linear layer followed by a linear projection to compute a score for each symbol in target vocabulary $V$. It then picks target symbol $z_t$ with the highest score:
\begin{align}
    generate(x) = \arg \max_{i=1}^{|V|} {z_i} \\
    z = W_{proj} \tanh(W_{out}^T x + b_{out}) + b_{proj}
\end{align}
with $W_{proj}\in\real^{|V|\times l},b_{proj}\in\real^{|V|}$, $W_{out}\in\real^{l\times(m'+2m+k)}$,
$b_{out}\in\real^l$, where $l$ is the output layer size.

\section{Experiments}
\label{exp}

\subsection{Model Settings}

Speech files were preprocessed using Yaafe \cite{mathieu2010}, to extract 40 MFCC features and frame energy for each frame with a step size of 10 ms and window size of 40 ms, following \cite{Chan2016,Berard2016}. We tokenize and lowercase all the text, and normalize the punctuation, with the Moses scripts\footnote{\url{http://www.statmt.org/moses/}}. For BTEC, the same preprocessing as \cite{Berard2016} is applied.
Character-level vocabularies for LibriSpeech are of size $46$ for English (transcriptions) and $167$ for French (translation). The decoder outputs are always at the character-level (for AST, MT and ASR).
For the MT task, the LibriSpeech English (source) side is preprocessed into subword units \cite{Sennrich2016}. We limit the number of merge operations to $30k$, which gives a vocabulary of size $27k$. The MT encoder for BTEC takes entire words as input.

Our BTEC models use an LSTM size of $m=m'=256$, while the LibriSpeech models use a cell size of $512$, except for the speech encoder layers which use a cell size of $m=256$ in each direction.  We use character embeddings of size $k=64$ for BTEC, and $k=128$ for LibriSpeech.
The MT encoders are more shallow, with a single bidirectional layer.
The source embedding sizes for words (BTEC) and subwords (LibriSpeech) are respectively $128$ and $256$.

The input layers in the speech encoders have a size of $256$ for the first layer and $n'=128$ for the second.
The LibriSpeech decoders use an output layer size of $l=512$. For BTEC, we do not use any non-linear output layer, as we found that this led to overfitting.

\subsection{Training settings}

We train our models with Adam \cite{Kingma2015}, with a learning rate of $0.001$, and a mini-batch size of 64 for BTEC, and 32 for LibriSpeech (because of memory constraints).
We use variational dropout \cite{Kingma2015b}, i.e., the same dropout mask is applied to all elements in a batch at all time steps, with a rate of $0.2$ for LibriSpeech and $0.4$ for BTEC. In the MT tasks, we also drop source and target symbols at random, with probability $0.2$. Dropout is not applied on recurrent connections \cite{Zaremba2014}.

We train all our models on \emph{LibriSpeech train} augmented with the Google Translate references, i.e., the source side of the corpus (speech) is duplicated, and the target side (translations) is a concatenation of the aligned references with the Google Translate references.
Because of GPU memory limits, we set the maximum length to $1400$ frames for LibriSpeech input, and $300$ characters for its output. This covers about $90\%$ of the training corpus. Longer sequences are kept but truncated to the maximum size. We evaluate our models on the dev set every 1000 mini-batch updates using BLEU for AST and MT, and WER for ASR, and keep the best performing checkpoint for final evaluation on the test set.

Our models are implemented with TensorFlow~\cite{Abadi2015} as part of the LIG-CRIStAL NMT toolkit\footnote{The toolkit and configuration files are available at: \url{https://github.com/eske/seq2seq}}.

\subsection{Results}

Table \ref{tab:BTEC} presents the results for the ASR and MT tasks on BTEC and LibriSpeech. The MT task (and by extension the AST task) on LibriSpeech (translating novels) looks particularly challenging, as we observe BLEU scores around 20\%\footnote{Google Translate is also scored as a topline (22.2\%).}.

\begin{table}[h]
    \centering
    \begin{tabular}{|c|c|c|c|}
        \hline
        & Model & ASR (WER $\downarrow$) & MT (BLEU $\uparrow$) \\
        \hline
        \multirow{3}{*}{\rotatebox{90}{BTEC}} & greedy & 14.9 & 47.4\\
        & beam-search & 13.8 & 49.2\\
        & ensemble & \textbf{11.3} & \textbf{50.7} \\
        \hline
        \hline
        \multirow{4}{*}{\rotatebox{90}{\footnotesize LibriSpeech}} & greedy & 19.9 & 19.2 \\
        & beam-search & 17.9 & 18.8 \\
        & ensemble & \textbf{15.1} & 19.3 \\
        & Google Translate & \cellcolor{black!20} & \textbf{22.2} \\
        \hline
    \end{tabular}
    \caption{MT and ASR results on test set for \emph{BTEC} and \emph{Augmented LibriSpeech}. We use a beam size of 8, and ensembles of 2 models trained from scratch.}
    \label{tab:BTEC}
\end{table}

\begin{table}[h]
    \centering
    \begin{tabular}{|c|c|c|c|c|}
         \hline
                    & greedy & beam & ensemble & params \\
                    \cline{2-4}
                    & \multicolumn{3}{c|}{Test BLEU} & (million) \\
         \hline
         Baseline \cite{Berard2016} & 29.1 & 31.3 & 37.9$\dagger$ & 10.4 \\
         \hline
         Cascaded & 38.9 & 40.7 & \textbf{43.8} & 7.9 + 3.4 \\
         \hline
         End-to-End & 31.3 & 33.7 & \cellcolor{black!20} & \multirow{3}{*}{6.7} \\
         \cline{1-4}
         Pre-trained & 33.7 & 36.3 & \multirow{2}{*}{\textbf{40.4}} & \\
         \cline{1-3}
         Multi-task & 35.1 & 37.6 & & \\
         \hline
    \end{tabular}
     \caption{Results of the AST task on \emph{BTEC test}.
    $\dagger$ was obtained with an ensemble of 5 models, while we use ensembles of 2 models. The non-cascaded ensemble combines the \emph{pre-trained} and \emph{multi-task} models. Contrary to \cite{Berard2016}, we only present mono-reference results.}

    \label{tab:BTEC-AST}
\end{table}

\captionsetup[figure]{aboveskip=0pt}
\captionsetup[figure]{belowskip=0pt}

\begin{figure}
    \centering
    \includegraphics[width=0.45\textwidth]{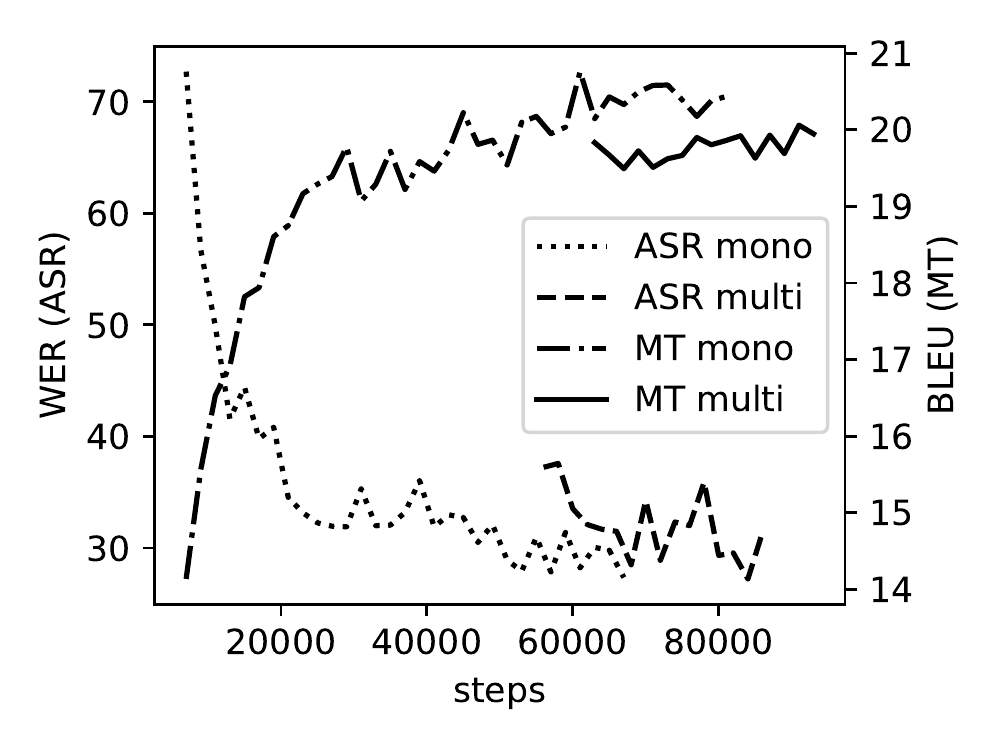}
    \caption{Augmented LibriSpeech Dev BLEU scores for the MT task, and WER scores for the ASR task, with the initial (mono-task) models, and when multi-task training picks up.}
    \label{fig:multitask_BLEU}
\end{figure}

For Automatic Speech Translation (AST), we try four settings. The \emph{cascaded} model combines both the ASR and MT models (as a pipeline). The \emph{end-to-end} model (described in section~\ref{models}) does not make any use of source language transcripts. The \emph{pre-trained} model is identical to \emph{end-to-end}, but its encoder and decoder are initialized with our ASR and MT models. The \emph{multi-task} model is also pre-trained, but continues training for all tasks, by alternating updates like \cite{Luong2016}, with 60\% of updates for AST and 20\% for ASR and MT.

Table~\ref{tab:BTEC-AST} and~\ref{tab:LIBR-AST} present the results for the end-to-end AST task on BTEC and LibriSpeech. On both corpora, we show that: (1) it is possible to train compact end-to-end AST models with a performance close to cascaded models; (2) pre-training and multi-task learning\footnote{if source transcriptions are available at training time} improve AST performance; (3) contrary to \cite{Weiss2017}, in both BTEC and LibriSpeech settings, best AST performance is observed when a symbolic sequence of symbols in the source language is used as an intermediary representation during the speech translation process (cascaded system); (4) finally, the AST results presented on LibriSpeech demonstrate that our augmented corpus is useful, although challenging, to benchmark end-to-end AST systems on real speech at a large scale.
We hope that the baseline we established on Augmented LibriSpeech will be challenged in the future.

The large improvements on MT and AST on the BTEC corpus, compared to~\cite{Berard2016} are mostly due to our use of a better decoder, which outputs characters instead of words.

\begin{table}[]
    \centering
    \begin{tabular}{|c|c|c|c|c|}
         \hline
                    & greedy & beam & ensemble & params \\
                    \cline{2-4}
                    & \multicolumn{3}{c|}{Test BLEU} & (million) \\
         \hline
         Cascaded   & 14.6   & 14.6 & \textbf{15.8}         & 6.3 + 15.9 \\
         \hline
         End-to-End & 12.3 & 12.9 & \multirow{3}{*}{\textbf{15.5$\dagger$}} & \multirow{3}{*}{9.4} \\
         \cline{1-3}
         Pre-trained & 12.6 & 13.3 & & \\
         \cline{1-3}
         Multi-task  & 12.6 & 13.4 & & \\
         \hline
    \end{tabular}
   \caption{AST results on \emph{Augmented LibriSpeech test}. $\dagger$ combines the end-to-end, pre-trained and multi-task models.}
    \label{tab:LIBR-AST}
\end{table}

\begin{figure}
    \centering
    \includegraphics[width=0.4\textwidth]{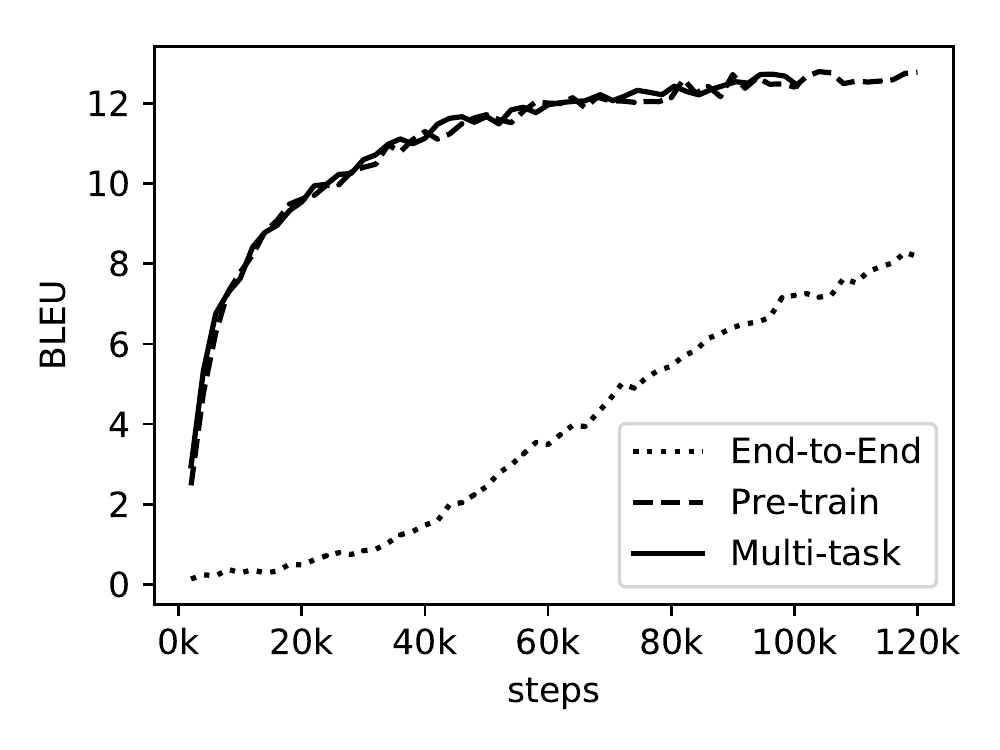}
   \caption{Dev BLEU scores on 3 models for end-to-end AST of audiobooks. Best scores on the dev set for the end-to-end (mono-task), pre-train and multi-task models were achieved at steps 369k, 129k and 95k.}
    \label{fig:AST_BLEU}
\end{figure}

\vspace{-.1cm}
\subsection{Analysis}

Figure~\ref{fig:multitask_BLEU} shows the evolution of BLEU and WER scores for MT and ASR tasks with single models, and when we continue training them as part of a multi-task model. The multi-task procedure does more updates on AST, which explains the degraded results, but we observe that the speech encoder and text decoder are still able to generalize well to other tasks.

Figure~\ref{fig:AST_BLEU} shows the evolution of dev BLEU scores for our three AST models on LibriSpeech. We see that pre-training helps the model converge much faster. Eventually, the End-to-End system reaches a similarly good solution, but after three times as many updates. Multi-Task training does not seem to be helpful when combined with pre-training.

\section{Conclusion}
\label{concl}

We present baseline results on End-to-End Automatic Speech Translation on
a new speech translation corpus of audiobooks, and on a synthetic corpus extracted from BTEC (follow-up to \cite{Berard2016}). We show that, while cascading two neural models for ASR and MT gives the best results, end-to-end methods that incorporate the source language transcript come close in performance.

\vfill\pagebreak

\bibliographystyle{IEEEbib}
\bibliography{main}

\begin{thebibliography}{10}

\bibitem{Duong2016}
Long Duong, Antonios Anastasopoulos, David Chiang, Steven Bird, and Trevor
  Cohn,
\newblock ``{An Attentional Model for Speech Translation Without
  Transcription},''
\newblock in {\em NAACL-HLT}, 2016.

\bibitem{Berard2016}
Alexandre B{\'{e}}rard, Olivier Pietquin, Laurent Besacier, and Christophe
  Servan,
\newblock ``{Listen and Translate: A Proof of Concept for End-to-End
  Speech-to-Text Translation},''
\newblock {\em NIPS 2016 End-to-end Learning for Speech and Audio Processing
  Workshop}, 2016.

\bibitem{Weiss2017}
Ron~J. Weiss, Jan Chorowski, Navdeep Jaitly, Yonghui Wu, and Zhifeng Chen,
\newblock ``{Sequence-to-Sequence Models Can Directly Transcribe Foreign
  Speech},''
\newblock in {\em Interspeech}, 2017.

\bibitem{Panayotov2015}
Vassil Panayotov, Guoguo Chen, Daniel Povey, and Sanjeev Khudanpur,
\newblock ``{Librispeech: an ASR corpus based on public domain audio books},''
\newblock in {\em ICASSP}, 2015.

\bibitem{Alican2018}
Ali~Can Kocabiyikoglu, Laurent Besacier, and Olivier Kraif,
\newblock ``{Augmenting Librispeech with French Translations: A Multimodal
  Corpus for Direct Speech Translation Evaluation},''
\newblock in {\em LREC}, 2018.

\bibitem{Adda2016}
Gilles Adda, Sebastian St\"{u}cker, Martine Adda-Decker, Odette Ambouroue,
  Laurent Besacier, David Blachon, H\'el\`ene Bonneau-Maynard, Pierre Godard,
  Fatima Hamlaoui, Dmitri Idiatov, Guy-No\"{e}l Kouarata, Lori Lamel,
  Emmanuel-Moselly Makasso, Annie Rialland, Mark Van~de Velde, Fran\c{c}ois
  Yvon, and Sabine Zerbian,
\newblock ``{Breaking the Unwritten Language Barrier: The Bulb Project},''
\newblock in {\em Proceedings of SLTU (Spoken Language Technologies for
  Under-Resourced Languages)}, 2016.

\bibitem{Anastasopoulos2017}
Antonios Anastasopoulos and David Chiang,
\newblock ``A case study on using speech-to-translation alignments for language
  documentation,''
\newblock in {\em Proceedings of the 2nd Workshop on the Use of Computational
  Methods in the Study of Endangered Languages}, 2017.

\bibitem{Post2013}
Matt Post, Gaurav Kumar, Adam Lopez, Damianos Karakos, Chris Callison-Burch,
  and Sanjeev Khudanpur,
\newblock ``{Improved Speech-to-Text Translation with the Fisher and Callhome
  Spanish-English Speech Translation Corpus},''
\newblock in {\em IWSLT}, 2013.

\bibitem{Bahdanau2015}
Dzmitry Bahdanau, Kyunghyun Cho, and Yoshua Bengio,
\newblock ``{Neural Machine Translation by Jointly Learning to Align and
  Translate},''
\newblock in {\em ICLR}, 2015.

\bibitem{Chorowski2015}
Jan Chorowski, Dzmitry Bahdanau, Dmitriy Serdyuk, Kyunghyun Cho, and Yoshua
  Bengio,
\newblock ``{Attention-Based Models for Speech Recognition},''
\newblock in {\em NIPS}, 2015.

\bibitem{Bahdanau2016}
Dzmitry Bahdanau, Jan Chorowski, Dmitriy Serdyuk, Philemon Brakel, and Yoshua
  Bengio,
\newblock ``{End-to-End Attention-based Large Vocabulary Speech Recognition},''
\newblock in {\em ICASSP}, 2016.

\bibitem{Sennrich2017}
Rico Sennrich, Orhan Firat, Kyunghyun Cho, Alexandra Birch, Barry Haddow,
  Julian Hitschler, Marcin Junczys-Dowmunt, Samuel Laeubli, Antonio Valerio,
  Antonio Valerio~Miceli Barone, Jozef Mokry, and Maria Nadejde,
\newblock ``{Nematus: a Toolkit for Neural Machine Translation},''
\newblock in {\em EACL}, 2017.

\bibitem{mathieu2010}
Benoit Mathieu, Slim Essid, Thomas Fillon, Jacques Prado, and Ga{\"e}l Richard,
\newblock ``{YAAFE, an Easy to Use and Efficient Audio Feature Extraction
  Software},''
\newblock in {\em ISMIR (International Society of Music Information
  Retrieval)}, 2010.

\bibitem{Chan2016}
William Chan, Navdeep Jaitly, Quoc~V. Le, and Oriol Vinyals,
\newblock ``{Listen, Attend and Spell},''
\newblock in {\em ICASSP}, 2016.

\bibitem{Sennrich2016}
Rico Sennrich, Barry Haddow, and Alexandra Birch,
\newblock ``{Neural Machine Translation of Rare Words with Subword Units},''
\newblock in {\em ACL}, 2016.

\bibitem{Kingma2015}
Diederik Kingma and Jimmy Ba,
\newblock ``{Adam: A Method for Stochastic Optimization},''
\newblock in {\em ICLR}, 2015.

\bibitem{Kingma2015b}
Diederik~P Kingma, Tim Salimans, and Max Welling,
\newblock ``Variational dropout and the local reparameterization trick,''
\newblock in {\em NIPS}, 2015.

\bibitem{Zaremba2014}
Wojciech Zaremba, Ilya Sutskever, and Oriol Vinyals,
\newblock ``{Recurrent Neural Network Regularization},''
\newblock in {\em ICLR}, 2014.

\bibitem{Abadi2015}
Martin Abadi, Ashish Agarwal, Paul Barham, Eugene Brevdo, Zhifeng Chen, Craig
  Citro, Greg Corrado, Andy Davis, Jeffrey Dean, Matthieu Devin, Sanjay
  Ghemawat, Ian Goodfellow, Andrew Harp, Geoffrey Irving, Michael Isard,
  Yangqing Jia, Lukasz Kaiser, Manjunath Kudlur, Josh Levenberg, Dan Man, Rajat
  Monga, Sherry Moore, Derek Murray, Jon Shlens, Benoit Steiner, Ilya
  Sutskever, Paul Tucker, Vincent Vanhoucke, Vijay Vasudevan, Oriol Vinyals,
  Pete Warden, Martin Wicke, Yuan Yu, and Xiaoqiang Zheng,
\newblock ``{TensorFlow: Large-Scale Machine Learning on Heterogeneous
  Distributed Systems},''
\newblock {\em arXiv}, 2015.

\bibitem{Luong2016}
Minh-Thang Luong, Quoc~V Le, Ilya Sutskever, Oriol Vinyals, and {\L}ukasz
  Kaiser,
\newblock ``{Multi-task Sequence to Sequence Learning},''
\newblock in {\em ICLR}, 2016.

\end{thebibliography}

\end{document}